\documentclass[12pt]{article}

\usepackage[margin=1in]{geometry}

\usepackage[utf8]{inputenc} 
\usepackage{url}            

\usepackage{graphicx}

\usepackage{longtable}

\usepackage{booktabs}

\usepackage{scalefnt}

\usepackage{amsmath,amssymb,amsfonts}
\usepackage{subcaption}
\usepackage{algorithm}
\usepackage{algorithmic}
\usepackage{sidecap}
\usepackage{arydshln}

\newcommand{\argmax}{\mathop{\rm arg~max}\limits}
\newcommand{\argmin}{\mathop{\rm arg~min}\limits}

\title{\bf A Simple Heuristic for\\ Bayesian Optimization with A Low Budget}

\author{
  Masahiro Nomura \\
  CyberAgent, Inc. \\
  \texttt{nomura\_masahiro@cyberagent.co.jp} \\
   \and
 Kenshi Abe \\
  CyberAgent, Inc. \\
  \texttt{abe\_kenshi@cyberagent.co.jp} \\
}

\date{}

\begin{document}
\maketitle

\begin{abstract}
The aim of black-box optimization is to optimize an objective function within the constraints of a given evaluation budget.
In this problem, it is generally assumed that the computational cost for evaluating a point is large; thus, it is important to search efficiently with as low budget as possible.
Bayesian optimization is an efficient method for black-box optimization and provides exploration-exploitation trade-off by constructing a surrogate model that considers uncertainty of the objective function.
However, because Bayesian optimization should construct the surrogate model for the entire search space, it does not exhibit good performance when points are not sampled sufficiently.
In this study, we develop a heuristic method refining the search space for Bayesian optimization when the available evaluation budget is low.
The proposed method refines a promising region by dividing the original region so that Bayesian optimization can be executed with the promising region as the initial search space.
We confirm that Bayesian optimization with the proposed method outperforms Bayesian optimization alone and shows equal or better performance to two search-space division algorithms through experiments on the benchmark functions and the hyperparameter optimization of machine learning algorithms.
\end{abstract}


\section{Introduction}

Black-box optimization is the problem of optimizing an objective function $f : \mathcal{X} \mapsto \mathbb{R}$ within the constraints of a given evaluation budget $B$.
In other words, the objective is to obtain a point ${\bf x} \in \mathcal{X}$ with the lowest possible evaluation value $f({\bf x})$ within the $B$ function evaluations.
In black-box optimization, no algebraic representation of $f$ is given, and no gradient information is available.
Black-box optimization involves problems such as hyperparameter optimization of machine learning algorithms~\cite{NIPS2011_4443,EggFeuBerSnoHooHutLey13,ilievski2017efficient,Golovin:2017:GVS:3097983.3098043}, parameter tuning of agent-based simulations~\cite{Yang2009AgentBasedSO}, and aircraft design~\cite{10.1007/978-3-540-31880-4_43}.

In black-box optimization, it is generally assumed that the computational cost for evaluating the point is large; thus it is important to search efficiently with as low budgets as possible.
For example, it is reported that the experiment of hyperparameter optimization of Online LDA takes about 12 days for 50 evaluations~\cite{NIPS2012_4522}.
The performance of deep neural networks (DNN) is known to be very sensitive to hyperparameters, and it has been actively studied in recent years~\cite{Bergstra:2012:RSH:2188385.2188395,Domhan:2015:SUA:2832581.2832731,ilievski2017efficient,DBLP:journals/jmlr/LiJDRT17,DBLP:conf/aistats/KleinFBHH17,pmlr-v80-falkner18a}.
Because the experiment of DNN also takes a long time to learn corresponding to one hyperparameter, a lower evaluation budget can be used for hyperparameter optimization.

Bayesian optimization is an efficient method for black-box optimization. 
Bayesian optimization is executed by repeating the following steps:
(1) Based on the data observed thus far, it constructs a surrogate model that considers the uncertainty of the objective function.
(2) It calculates the acquisition function to determine the point to be evaluated next by using the surrogate model constructed in step (1).
(3) By maximizing the acquisition function, it determines the point to be evaluated next.
(4) It then updates the surrogate model based on the newly obtained data, then returns to step (2).

However, Bayesian optimization, which constructs a surrogate model for the entire search space, can show bad performance in the low budget setting, because an optimization method cannot sample points sufficiently.
In the low budget setting, we believe that a search should be performed locally; however, in Bayesian optimization, to estimate uncertainty in the search space by surrogate model, the points are sampled for the search space globally.
Therefore, the lack of local search degrades the performance of Bayesian optimization.
If there is no prior knowledge of the problem, the search space tends to be widely defined.
When the search space is widely defined, the search will be performed more globally, degrading performance.

In this study, we develop a heuristic method that refines the search space for Bayesian optimization when evaluation budget is low.
The proposed method performs division to reduce the volume of the search space. 
The proposed method makes it possible to perform Bayesian optimization within the local search space determined to be promising.
We confirm that Bayesian optimization with the proposed method outperforms Bayesian optimization alone (that is, Bayesian optimization without the proposed method) by the experiments on the six benchmark functions and the hyperparameter optimization of the three machine learning algorithms (multi-layer perceptron (MLP), convolutional neural network (CNN), LightGBM).
We also experiment with Simultaneous Optimistic Optimization (SOO)~\cite{NIPS2011_4304} and BaMSOO~\cite{pmlr-v33-wang14d}, which are search-space division algorithms, in order to confirm the validity of the refinement of the search space by the proposed method.

\section{Background}

\subsection{Bayesian Optimization}
\label{sec:bo}

Algorithm \ref{alg: bo} shows the algorithm of Bayesian optimization, which samples and evaluates the initial points (line 1),
constructs the surrogate model (line 3),
finds the next point to evaluate by optimizing the acquisition function (line 4),
evaluates the point selected and receives the evaluation value (line 5),
and updates the data (line 6).

The main components of Bayesian optimization are the surrogate model and the acquisition function.
In this section, we describe Bayesian optimization using a Gaussian process as the surrogate model and the expected improvement (EI) as the acquisition function.

\begin{algorithm}
\caption{Bayesian Optimization}
\label{alg: bo}
\begin{algorithmic}[1]
  \REQUIRE objective function $f$, search space $\mathcal{X}$, initial sample size $n$, surrogate model $\mathcal{M}$, acquisition function $\alpha({\bf x} \mid \mathcal{M})$
  \STATE sample and evaluate initial points: $\mathcal{D}_0 = \{({\bf x}_{0,1}, y_{0,1}), \cdots, ({\bf x}_{0, n}, y_{0, n})\}$
  \FOR{$t = 1,2,\cdots$}
      \STATE construct surrogate model $\mathcal{M}$ by $\mathcal{D}_{t-1}$
      \STATE find ${\bf x}_t$ by optimizing the acqusition function $\alpha$ : ${\bf x}_t = \argmax_{{\bf x} \in \mathcal{X}} \alpha({\bf x} \mid \mathcal{M})$
      \STATE evaluate ${\bf x}_t$ and receive: $y_t = f({\bf x}_t)$
      \STATE update the data $\mathcal{D}_t = \mathcal{D}_{t-1} \cup \{({\bf x}_t, y_t)\}$
  \ENDFOR
\end{algorithmic}
\end{algorithm}

\subsubsection{Gaussian Process}
A Gaussian process~\cite{Rasmussen:2005:GPM:1162254} is the probability distribution over the function space characterized by the mean function $\mu: \mathcal{X} \to \mathbb{R}$ and the covariance function $\sigma^2: \mathcal{X} \to \mathbb{R}_{\geq 0}$.
We assume that data set $\mathcal{D}_t = \{({\bf x}_1, y_1) \cdots, ({\bf x}_{t}, y_{t})\}$ and observations ${\bf y} = ( y_{1}, \cdots , y_t )$ are obtained.
The mean $\mu({\bf x}_{t+1})$ and variance $\sigma^2({\bf x}_{t+1})$ of the predicted distribution $f({\bf x}_{t+1}) \sim \mathcal{N}(\mu({\bf x}_{t+1}), \sigma^2({\bf x}_{t+1}))$ of a Gaussian process with respect to ${\bf x}_{t+1}$ can be calculated using the kernel function $k: \mathcal{X} \times \mathcal{X} \to \mathbb{R}_{\geq 0}$ as follows:
\begin{align}
\mu({\bf x}_{t+1}) & = {\bf k}^{\top} {\bf K}^{-1} {\bf y}, \\
\sigma^2({\bf x}_{t+1}) & = k({\bf x}_{t+1}, {\bf x}_{t+1}) - {\bf k}^{\top} {\bf K}^{-1} {\bf k}.
\end{align}
Here, 
\begin{align}
{\bf K} &= 
    \begin{bmatrix}
      k({\bf x}_{1}, {\bf x}_{1}) & \cdots & k({\bf x}_{1}, {\bf x}_{t}) \\
      \vdots & \ddots & \vdots\\
      k({\bf x}_{t}, {\bf x}_{1}) & \cdots & k({\bf x}_{t}, {\bf x}_{t})
    \end{bmatrix}, \\
\mathbf{k} &= \begin{bmatrix}k({\bf x}_{t+1}, {\bf x}_{1}) & \cdots & k({\bf x}_{t+1}, {\bf x}_{t})\end{bmatrix}^{\top}.
\end{align}
The squared exponential kernel (Equation (\ref{eq:se})) is one of the common kernel functions.
\begin{align}
\label{eq:se}
k_{\rm SE} ({\bf x}, {\bf x}^{\prime}) & = \sigma_f \exp \left(-\frac{r^2({\bf x}, {\bf x}^{\prime})}{2}\right), \\
\label{eq:kernel_r}
r^2({\bf x}, {\bf x}^{\prime}) & = \frac{\|{\bf x} - {\bf x}^{\prime}\|^2}{l}.
\end{align}
Here, $\sigma_f$ is a parameter that adjusts the scale of the whole kernel function, and $l$ is a parameter of sensitivity to the difference between the two inputs ${\bf x}, {\bf x}^{\prime}$.

\subsubsection{Expected Improvement}
The EI \cite{jones1998efficient} is a typical acquisition function in Bayesian optimization, and it represents the expectation value of the improvement amount for the best evaluation value of the candidate point.
Let the best evaluation value to be $f_{\min}$, EI for the point ${\bf x}$ is calculated as follow:
\begin{equation}
\label{eq: ei_origin}
\alpha_{\rm EI}({\bf x} \mid \mathcal{D}) = \mathbb{E}[\max\{f_{\min} - f({\bf x}), 0\} \mid {\bf x}, \mathcal{D}].
\end{equation}
When we assume that the objective function follows a Gaussian process, Equation (\ref{eq: ei_origin}) can be calculated analytically as follows:
\begin{equation}
\begin{split}
\alpha_{\rm EI}({\bf x} \mid \mathcal{D}) &=\begin{cases}
    \sigma(x) \cdot \phi (Z) + (f_{\min} - \mu(x)) \cdot \Phi (Z) & (\sigma({\bf x}) > 0) \\
    0 & (\sigma({\bf x}) = 0),
\end{cases} \\
Z &= \frac{f_{\min} - \mu({\bf x})}{\sigma({\bf x})}.
\end{split}
\end{equation}
Here, $\Phi$ and $\phi$ are the cumulative distribution function and probability density function of the standard normal distribution, respectively.

\subsection{Related Work}
\subsubsection{Bayesian optimization}
In Bayesian optimization, the design of surrogate models and acquisition functions are actively studied.
The tree-structured Parzen Estimator (TPE) algorithm \cite{NIPS2011_4443,pmlr-v28-bergstra13}, Sequential Model-based Algorithm Configuration (SMAC) \cite{hutter2011sequential} and Spearmint \cite{NIPS2012_4522} are known as powerful Bayesian optimization methods.
The TPE algorithm, SMAC, and Spearmint use a tree-structured parzen estimator, a random forest, and a Gaussian process as the surrogate model, respectively.
The popular acquisition functions in Bayesian optimization include the EI \cite{jones1998efficient}, probability of improvement \cite{article_pi}, upper confidence bound (UCB) \cite{Srinivas:2010:GPO:3104322.3104451}, mutual information (MI) \cite{pmlr-v32-contal14}, and knowledge gradient (KG) \cite{Frazier2009TheKP}.

However, there are few studies focusing on search spaces in Bayesian optimization.
A prominent problem in Bayesian optimization is the boundary problem~\cite{SwerskyDThesis} that points sampled concentrate near the boundary of the search space.
Oh et al. addressed this boundary problem by transforming the ball geometry of the search space using cylindrical transformation~\cite{pmlr-v80-oh18a}.
Wistuba et al. proposed using the previous experimental results to prune the search space of hyperparameters where there seems to be no good point \cite{10.1007/978-3-319-23525-7_7}.
In contrast to Wistuba's study, we propose a method to refine the search space without prior knowledge.
Nguyen et al. dynamically expanded the search space to cope with cases where the search space specified in advance does not contain a good point~\cite{inproceedings}.
In contrast to Nguyen's study, we focus on refining the search space rather than expanding.

\subsubsection{Search-Space Division Algorithm}

The proposed method is similar to methods such as Simultaneous Optimistic Optimization (SOO)~\cite{NIPS2011_4304} and BaMSOO~\cite{pmlr-v33-wang14d} in that it focuses on the division of the search space.
SOO is an algorithm that generalizes the DIRECT algorithm~\cite{Jones1993}, which is a Lipschitz optimization method, and the search space is expressed as a tree structure and the search is performed using hierarchical division.
BaMSOO is a method that makes auxiliary optimization of acquisition functions unnecessary by combining SOO with Gaussian process.
Wang et al. reported that BaMSOO shows better performance than SOO in experiments on some benchmark functions~\cite{pmlr-v33-wang14d}.
In the proposed method and search-division algorithms, SOO and BaMSOO, the motivation for optimization is different;
the proposed method divides the search space to identify a promising initial region for Bayesian optimization, while the search-division algorithms divide the search space to identify a good solution.

\section{Proposed Method}
In Bayesian optimization, there are many tasks with a low available evaluation budget.
For example, in hyperparameter optimization of machine learning algorithms, budget would be limited in terms of computing resources and time.
In this study, we focus on Bayesian optimization when there is not enough evaluation budget available.

Nguyen et al. state that Bayesian opitmization using a Gaussian process as the surrogate model and UCB as the acquisition function has the following relationships between the volume of a search space and the cumulative regret (the sum of differences between the optimum value and the evaluation value at each time)~\cite{inproceedings}.
(i) A larger space will have larger (worse) regret bound.
(ii) A low evaluation budget will make the difference in the regrets more significant.
Nguyen et al. give above description for cumulative regret~\cite{inproceedings}, but converting it to simple regret is straightforward, such as  ~\cite{NIPS2016_6118}.
We therefore believe that in the low budget setting, making the search space smaller is also important in terms of the regret for Bayesian optimization in general.

In this study, we try to improve the performance of Bayesian optimization with the low budget setting by introducing a heuristic method that refines a given search space.
We assume that we have an arbitrary hypercube $\mathcal{X} \subseteq \mathbb{R}^d$ ($d$: the number of dimensions) as a search space.
Our method refines the search space by division, and outputs a region $\mathcal{X}_S (\subseteq \mathcal{X})$ considered to be promising.
As a result, Bayesian optimization can be executed with the refined search space $\mathcal{X}_S$ as the initial search space instead of the original search space $\mathcal{X}$.

\subsection{Integrating with Bayesian Optimization}
\label{sec:integrating_bo}
Algorithm \ref{alg:refine_frame} shows Bayesian optimization with the proposed method.
This method calculates the budget $B_{\rm ref}$ for refining the search space from the whole budget $B$ (line 1),
refines the promising search space (line 2),
performs optimization with the search space $\mathcal{X}_S$ refined in the line 2 as the initial search space (line 3).
We will describe ${\rm refine\_search\_space}(d, \mathcal{X}, B_{\rm ref})$ (line 2) in Section \ref{sec:proposal_algorithm}.

\begin{algorithm}
\caption{Bayesian optimization with the proposed method}
\label{alg:refine_frame}
\begin{algorithmic}[1]
    \REQUIRE budget $B$, search space $\mathcal{X}$, the number of dimension $d$, ratio for refining $\gamma \ (0 \leq \gamma \leq 1)$
    \STATE $B_{\rm ref} \gets \gamma \cdot B$
    \STATE $\mathcal{X}_{S} \gets {\rm refine\_search\_space}(d, \mathcal{X}, B_{\rm ref})$
    \STATE Bayesian optimization with the search space $\mathcal{X}_S$ until a budget reach $B$
\end{algorithmic}
\end{algorithm}

\begin{figure}[t]
\centering
    \includegraphics[width=70mm]{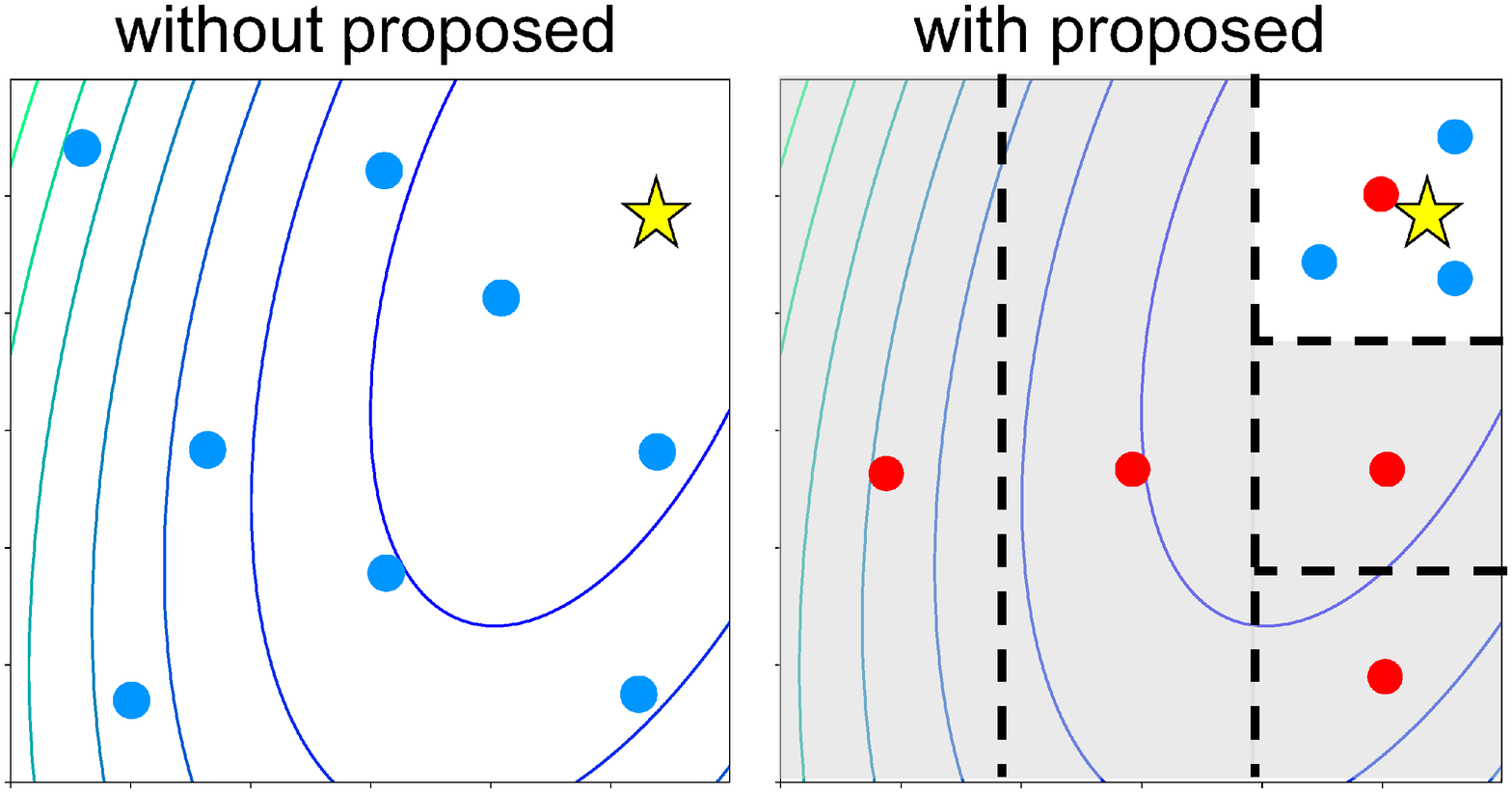}
    \caption{Conceptual design of Bayesian optimization with the proposed method. The figure on the left shows Bayesian optimization without the proposed method, and the figure on the right shows Bayesian optimization with the proposed method. The blue balls are the points sampled by Bayesian optimization and the red balls are the points sampled by the proposed method. The gray region on the right shows the discarded region by the proposed method.}
    \label{fig:basic_idea}
\end{figure}

Figure \ref{fig:basic_idea} shows a conceptual design of the proposed method.
In the Figure \ref{fig:basic_idea} on the right, Bayesian optimization is executed on the search space which has been refined by the proposed method.

\subsection{Refining the Search Space}

\subsubsection{Calculation of the Budget}
\label{sec:budget_ratio}
Corresponding to the whole budget $B$, we set the budget $B_{\rm ref}$ used for the proposed method to $B_{\rm ref} = \gamma \cdot B$ (in Algorithm~\ref{alg:refine_frame}, line 1).
We calculate $\gamma$ by $\gamma = 0.59 \cdot \exp(- 0.033 B / d)$ with respect to the number of dimension $d \ (\in \mathbb{N})$.
If the evaluation budget increases to infinity (that is, $B \to \infty$), there is no need for refining the search space (that is, $\gamma \to 0$).
We note that $B_{\rm ref}$ is maximum budget for the proposed method, not used necessarily in fact; $B_{\rm ref}$ is used for determining the division number.
We show the details about how $B_{\rm ref}$ is used in Section \ref{sec:division_number}.

\subsubsection{Algorithm}
\label{sec:proposal_algorithm}
\begin{figure}[t]
\centering
    \includegraphics[width=90mm]{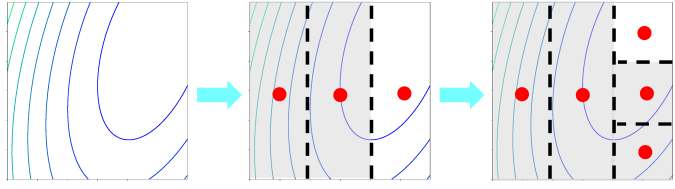}
    \caption{Refining the search space by the proposed method for the number of dimensions $d = 2$ and the division number $K = 3$. The proposed method divides the region corresponding to a certain dimension into $K$ pieces, and evaluates the center points. The proposed method leaves the region where the center point is the best, divides the region corresponding to another certain dimension again into $K$ pieces, and repeats this.}
    \label{fig:refine}
\end{figure}

The proposed method refines the promising region by dividing the region at equal intervals for each dimension.
Figure \ref{fig:refine} shows that refining the search space by the proposed method when the number of dimensions is $d = 2$ and the division number is $K = 3$.
The proposed method randomly selects a dimension without replacement, divides the region corresponding to the dimension into $K$ pieces, leaves only the region where the evaluation value of the center point of the divided region is the best.
The proposed method repeats this operation until the division of the regions corresponding to all dimensions is completed.

Algorithm \ref{alg:refine_search} shows the algorithm of refining the search space.
We denote the set of integers between $1$ and $n$ (including $1$ and $n$) by $[n]$ throughout the paper.
We describe how to set the division number $K$ (in Algorithm \ref{alg:refine_search}, line 1) in the next section.

\begin{algorithm}
\caption{${\rm refine\_search\_space}(d, \mathcal{X}, B_{\rm ref})$}
\label{alg:refine_search}
\begin{algorithmic}[1]
    \REQUIRE the number of dimension $d$, search space $\mathcal{X}$, budget for refining $B_{\rm ref}$
    \STATE $K = \argmax_{k \in \mathbb{N} \setminus 2 \mathbb{Z}} \{ B_k : B_k \leq B_{\rm ref} \}$
    \IF{$K \leq 1$}
      \STATE return $\mathcal{X}$ \COMMENT{need not divide in this case}
    \ENDIF
    \STATE initialization : $\mathcal{X}_S = \mathcal{X}$
    \FOR {$i = 1$ to $d$}
      \STATE randomly select an index $d_i$ without replacement from index set of dimensions
      \STATE divide $\mathcal{X}_S$ into $\{ X_{k} \}_{k=1}^K$ with respect to dimension $d_i$
      \STATE $k^{*} = \argmin_{k \in [K]} f(x_k)$, $x_k$ is a center point of $X_k$
      \STATE update the search space : $\mathcal{X}_S \gets X_{k^{*}}$
    \ENDFOR
    \STATE return $\mathcal{X}_S$
\end{algorithmic}
\end{algorithm}

\subsubsection{Division Number}
\label{sec:division_number}
We need to set the division number $K$ to adjust how much the search space is refined.
If we set the division number $K$ to an even number, the evaluation budget for refining is calculated by $B_K^{\rm even} = d K$.
However, when $K$ is an odd number, the evaluation budget for refining is calculated by $B_K^{\rm odd} = K + \sum_{l=1}^{d-1} (K - 1)$ because the center point of the search space refined before can be reused for the next evaluation.
Therefore, we set the division number $K$ to an odd number $K \in \mathbb{N} \setminus 2 \mathbb{Z}$ so that the evaluation budget for refining $B_K$ approaches $B_{\rm ref}$ most closely according to Equation (\ref{eq:K}).

\begin{equation}
  \label{eq:K}
  K = \argmax_{k \in \mathbb{N} \setminus 2 \mathbb{Z}} \{ B_k : B_k \leq B_{\rm ref} \}.
\end{equation}

\section{Experiments}
In this section, we assess the performance of the proposed method through the benchmark functions and the hyperparameter optimization of machine learning algorithms to confirm the effectiveness of the proposed method in the low budget setting.

\subsection{Baseline Methods}
We use GP-EI (Bayesian optimization using Gaussian process as the surrogate model and the EI as the acquisition function), TPE \cite{NIPS2011_4443} and SMAC \cite{hutter2011sequential} as the baseline methods of Bayesian optimization.
In this experiment, we refer to GP-EI with the proposed method as {\it Ref+GP-EI}.
Likewise, we refer to TPE and SMAC with the proposed method as {\it Ref+TPE} and {\it Ref+SMAC}, respectively.
We use the GPyOpt\footnote{https://github.com/SheffieldML/GPyOpt}, Hyperopt\footnote{https://github.com/hyperopt/hyperopt} and SMAC3\footnote{https://github.com/automl/SMAC3} library to obtain the results for GP-EI, TPE and SMAC, respectively.
We set the parameters of GP-EI, TPE, and SMAC to the default values of each library and use the center point of the search space as the initial starting point for SMAC.
We also experiment with SOO~\cite{NIPS2011_4304} and BaMSOO~\cite{pmlr-v33-wang14d}, which are search-space division algorithms, in order to confirm the validity of the refinement of the search space by the proposed method.

\subsection{Benchmark Functions}
In the first experiment, we assess the performance of the proposed method on the benchmark functions that are often used in black-box optimization.
Table \ref{tab:bench} shows the six benchmark functions used in this experiment.

\begin{table*}[tb]
\caption{Name, definition formula, number of dimension, and search space of the benchmark functions. The coefficients appearing in the Branin, Shekel, and Harmann functions are shown in \cite{Benchmark2014}.}
\label{tab:bench}
\begin{center}
  \begin{tabular}{l|c|c|c}
    \hline
    {\footnotesize Name} & {\footnotesize Definition} & {\footnotesize Dim $d$} & {\footnotesize Search Space $\mathcal{X}$} \\
    \hline \hline
    {\footnotesize Sphere} & {\footnotesize $f({\bf x}) = \sum_{i=1}^d x_i^2$} & {\footnotesize $5$} & {\footnotesize $[-5, 10]^{5}$} \\
    \hline
    {\footnotesize $k$-tablet $(k = \lfloor d/4 \rfloor)$} & {\footnotesize $f({\bf x}) = \sum_{i=1}^k x_i^2 + \sum_{i=k+1}^d (100 x_i)^2$} & {\footnotesize $5$} & {\footnotesize $[-5, 10]^{5}$} \\
    \hline
    {\footnotesize RosenbrockChain} & {\footnotesize $f({\bf x}) = \sum_{i=1}^{d-1}\left(100(x_{i+1} - x_i^2)^2 + (x_i - 1)^2\right)$} & {\footnotesize $5$} & {\footnotesize $[-5, 10]^{5}$} \\
    \hline
    {\footnotesize Branin} & {\footnotesize \begin{tabular}{c} $f({\bf x}) = a (x_2 - b x_1^2 + c x_1 - r)^2$\\ \hspace{10mm}$ + s (1 - t) {\rm cos}(x_1) + s$ \end{tabular} } & {\footnotesize $2$} & {\footnotesize $[-5, 10] \times [0, 15]$} \\
    \hline
    {\footnotesize Shekel $(m = 5)$} & {\footnotesize $f({\bf x}) = - \sum_{i=1}^5 \Bigl( \sum_{j=1}^4 (x_j - C_{ji})^2 + \beta_{i} \Bigr)^{-1}$} & {\footnotesize $4$} & {\footnotesize $[0, 10]^4$} \\
    \hline
    {\footnotesize Hartmann} & {\footnotesize $f({\bf x}) = - \sum_{i=1}^4 \alpha_i \exp \Bigl( - \sum_{j=1}^6 A_{ij} (x_j - P_{ij})^2 \Bigr)$} & {\footnotesize $6$} & {\footnotesize $[0, 1]^6$} \\
    \hline
  \end{tabular}
\end{center}
\end{table*}

\subsubsection{Experimental Setting}
We run 50 trials for each experiment, and we set the evaluation budget to $B = 10 \times d$ in each trial.
We assess the performance of each method using the mean and standard error of the best evaluation values in 50 trials.

In SOO and BaMSOO, we set the division numner $K = 3$, which is the same setting in ~\cite{pmlr-v28-valko13}.
For BaMSOO, we use the Mat$\acute{e}$rn $5/2$ kernel, which is one of the common kernel functions.
This equation is given by $k_{\rm M52} ({\bf x}, {\bf x}^{\prime}) = \sigma_f (1 + \sqrt{5 r^2({\bf x}, {\bf x}^{\prime})} + 5 r^2({\bf x}, {\bf x}^{\prime}) / 3) \exp (- \sqrt{5 r^2({\bf x}, {\bf x}^{\prime})}) ,$ where $r^2({\bf x}, {\bf x}^{\prime}) = \|{\bf x} - {\bf x}^{\prime}\|^2 / l$.
We set the initial hyperparameters to $\sigma_f = 1$ and $l = 0.25$ and update them by maximizing the data likelihood after each iteration.

\subsubsection{Results}
\begin{figure*}[tb]
\centering
  \begin{minipage}[t]{.47\textwidth}
    \centering
    \includegraphics[width=70mm]{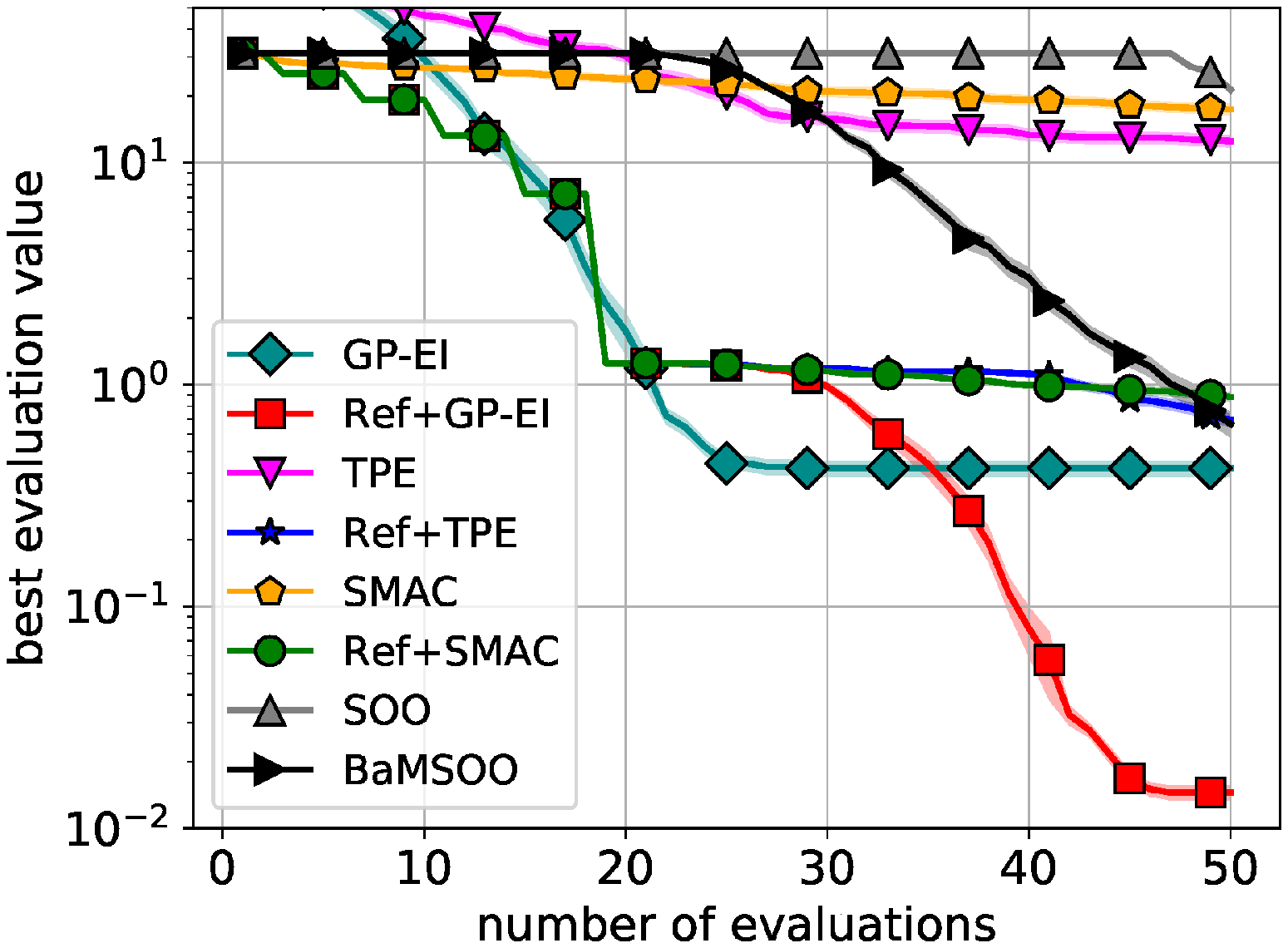}
    \subcaption{Sphere}
    \label{fig:mean1}
  \end{minipage}
  \begin{minipage}[t]{.47\textwidth}
    \centering
    \includegraphics[width=70mm]{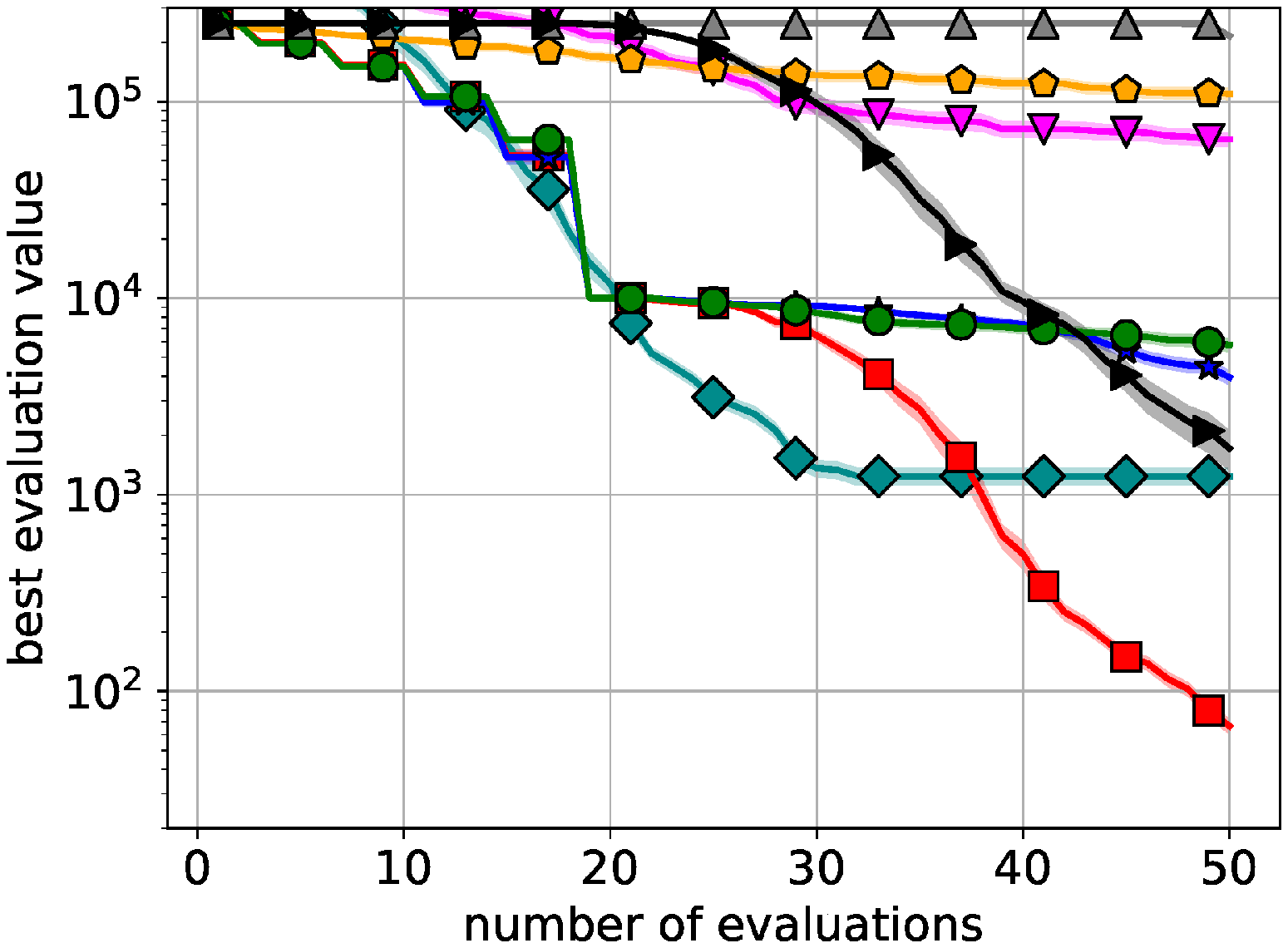}
    \subcaption{$k$-tablet $(k = d/4)$}
    \label{fig:mean1}
  \end{minipage}
  \begin{minipage}[t]{.47\textwidth}
    \centering
    \includegraphics[width=70mm]{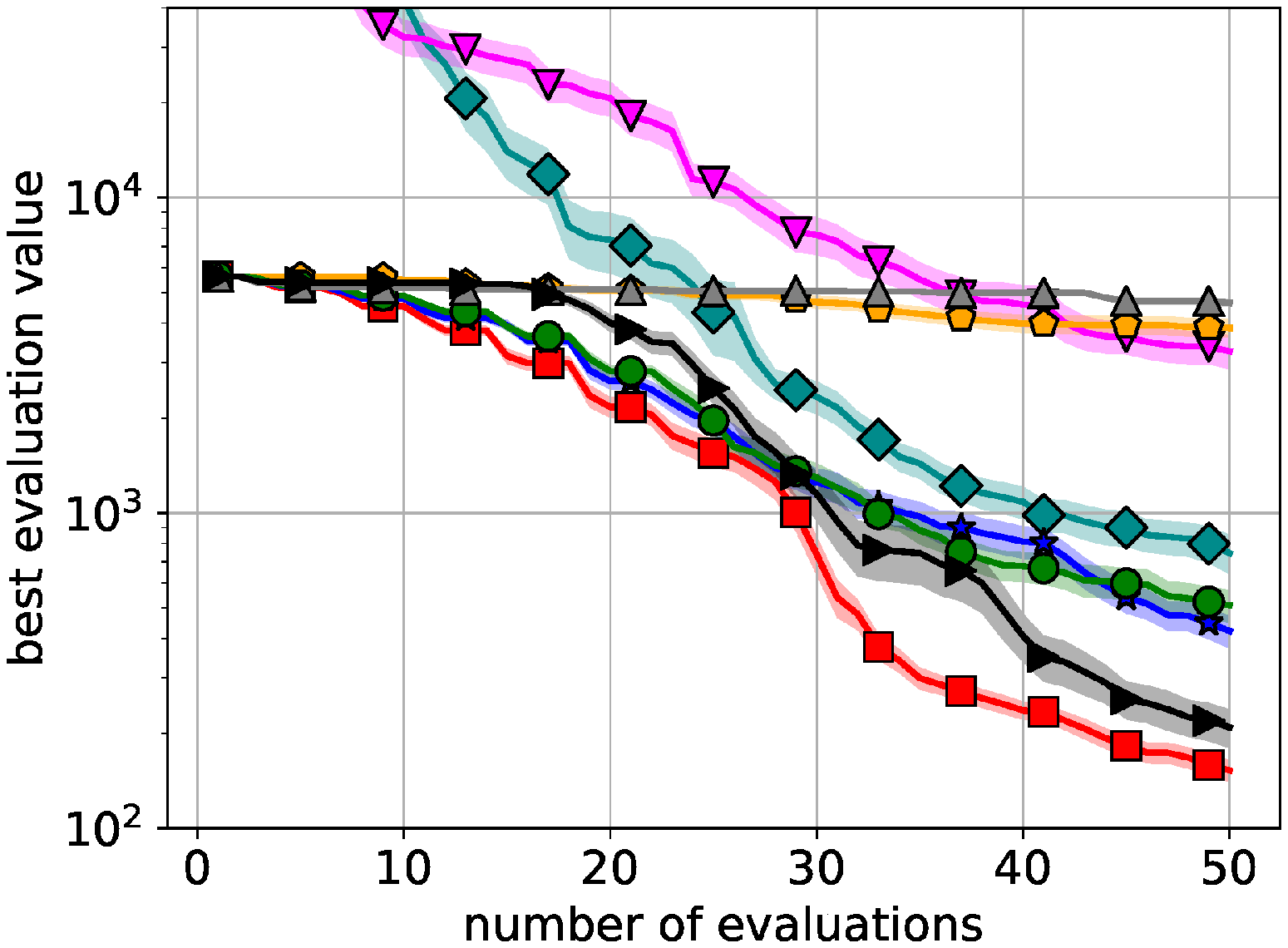}
    \subcaption{RosenbrockChain}
    \label{fig:mean1}
  \end{minipage}
  \begin{minipage}[t]{.47\textwidth}
    \centering
    \includegraphics[width=70mm]{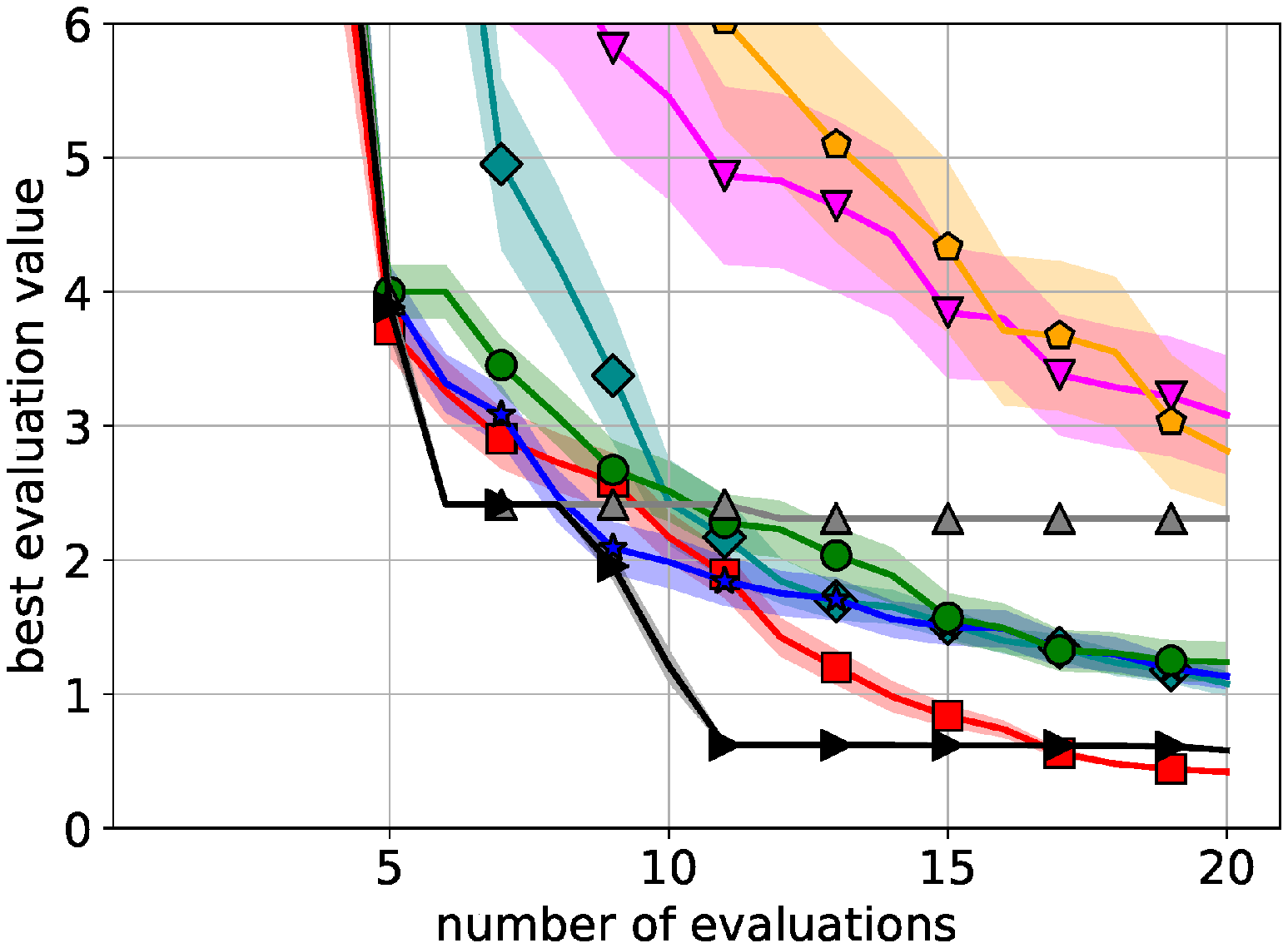}
    \subcaption{Branin}
    \label{fig:mean1}
  \end{minipage}
  \begin{minipage}[t]{.47\textwidth}
    \centering
    \includegraphics[width=70mm]{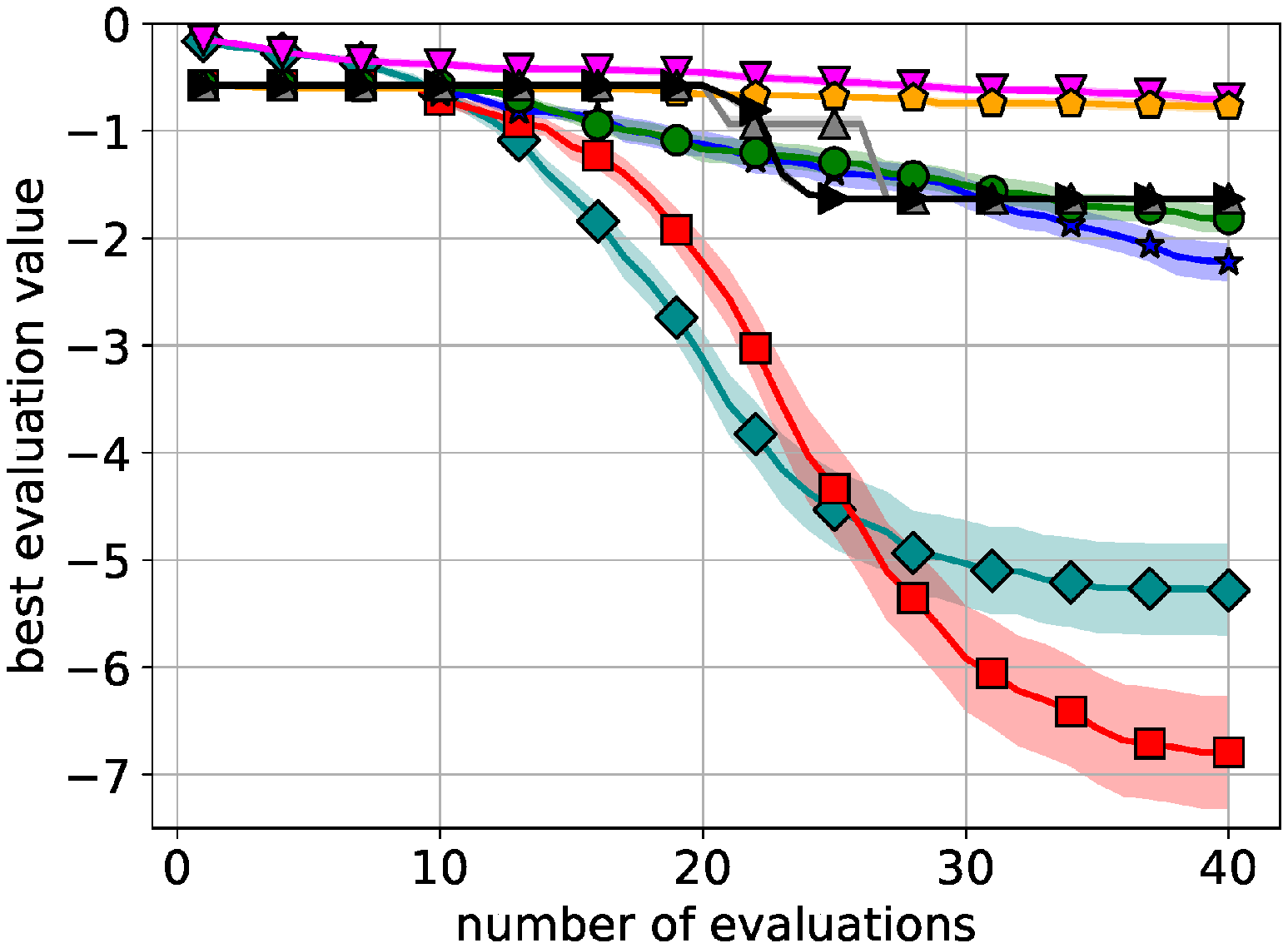}
    \subcaption{Shekel}
    \label{fig:mean1}
  \end{minipage}
  \begin{minipage}[t]{.47\textwidth}
    \centering
    \includegraphics[width=70mm]{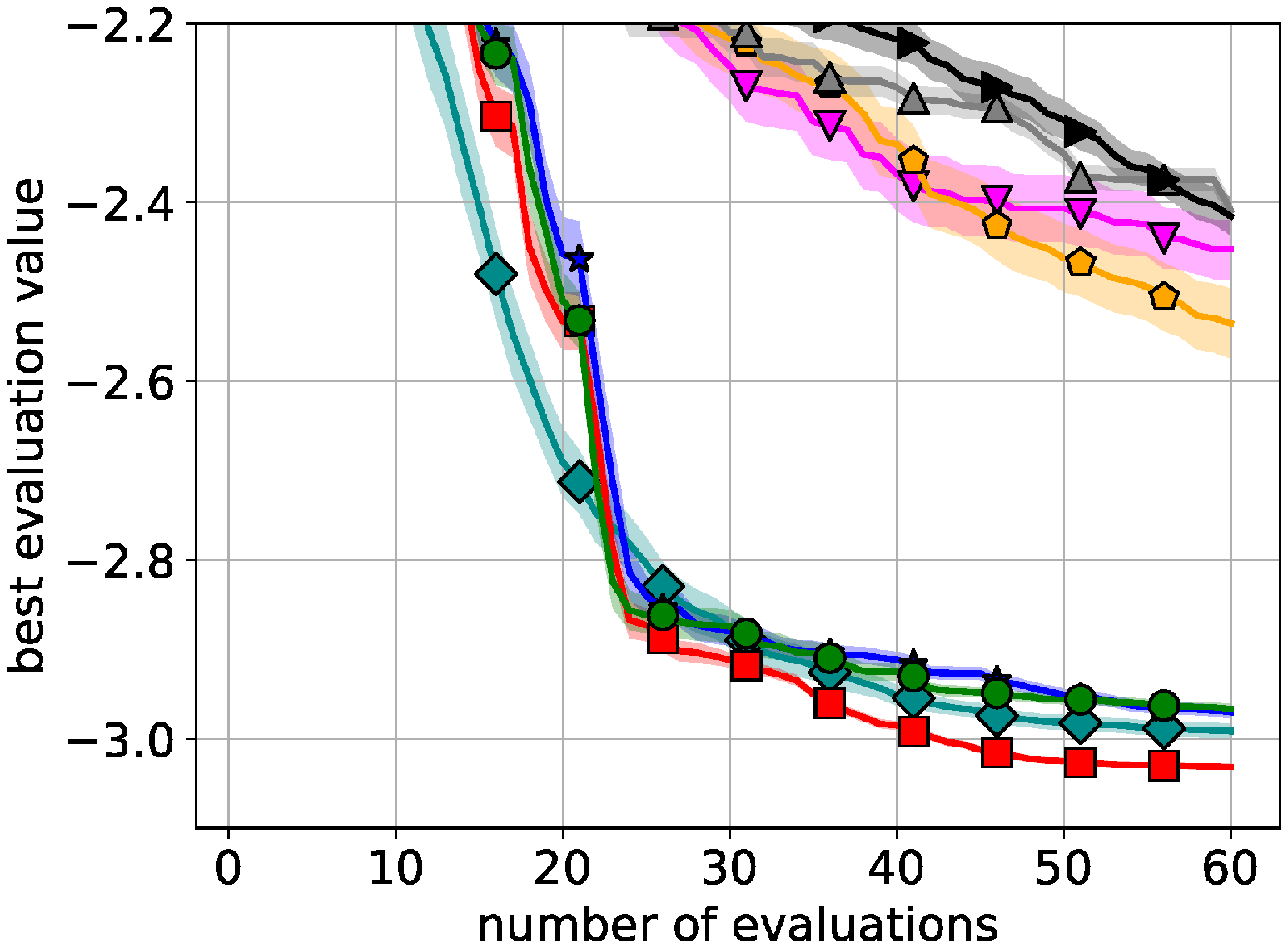}
    \subcaption{Hartmann}
    \label{fig:mean1}
  \end{minipage}
  \caption{Sequences of the mean and standard error of the best evaluation values on the benchmark functions. The x-axis denotes the number of evaluations and the y-axis denotes the mean and standard error of the best evaluation values (averaged over 50 trials).}
  \label{fig:exp_bench}
\end{figure*}

Figure \ref{fig:exp_bench} and Table \ref{tab:exp_bench} show the mean and standard error of the best evaluation values in 50 trials on the six benchmark functions.
Ref+GP-EI and BaMSOO show competitive performance in RosenbrockChain and Branin function, but Ref+GP-EI shows better performance than all the other methods in other benchmark functions.
Furthermore, Ref+GP-EI, Ref+TPE and Ref+SMAC outperform GP-EI, TPE and SMAC in all the benchmark functions, respectively.

\begin{figure*}[tb]
\centering
    \includegraphics[width=75mm]{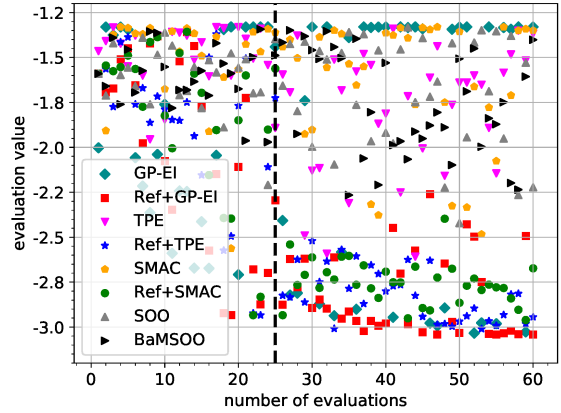}
    \caption{Typical behavior of each method on the Hartmann function. The x-axis denotes the number of evaluations and the y-axis denotes the evaluation value. The black dotted line represents the moment that the proposed method refines the search space.}
    \label{fig:discuss_typical_behavior}
\end{figure*}

Figure \ref{fig:discuss_typical_behavior} shows the typical behavior of each method on the Hartmann function.
Ref+GP-EI, Ref+TPE and Ref+SMAC sample many points with good evaluation values after refining the search space whereas other methods have not been able to sample points sufficiently with good evaluation values even at the end of the search.

\begin{table*}[tb]
\caption{Mean and standard error of the best evaluation values on the benchmark functions. The bold line shows the best mean in all the methods. The values in the $k$-tablet function and the RosenbrockChain function are multiplied by $10^3$ and $10^2$ with the original values, respectively. }
\label{tab:exp_bench}
\small
\begin{center}
  {\tabcolsep = 1mm
  \begin{tabular}{c||c|c|c|c|c|c|c|c}
    \hline
    Problem & GP-EI & Ref+GP-EI & TPE & Ref+TPE & SMAC & Ref+SMAC & SOO & BaMSOO \\
    \hline \hline
    Sphere & \begin{tabular}{c} $0.420$\\ $\pm 0.04$ \end{tabular} & \begin{tabular}{c} ${\bf 0.0145}$\\ $\pm 0.001$ \end{tabular} & \begin{tabular}{c} $12.5$\\ $\pm 0.8$ \end{tabular} & \begin{tabular}{c} $0.694$\\ $\pm 0.05$ \end{tabular} & \begin{tabular}{c} $17.4$\\ $\pm 1.0$ \end{tabular} & \begin{tabular}{c} $0.883$\\ $\pm 0.04$ \end{tabular} & \begin{tabular}{c} $21.4$\\ $\pm 0.3$ \end{tabular} & \begin{tabular}{c} $0.665$\\ $\pm 0.09$ \end{tabular} \\
    \hline
    $k$-tablet & \begin{tabular}{c} $1.24$\\ $\pm 1$ \end{tabular} & \begin{tabular}{c} ${\bf 0.0663}$\\ ${\pm 0.006}$ \end{tabular} & \begin{tabular}{c} $64$\\ $\pm 6$ \end{tabular} & \begin{tabular}{c} $3.95$\\ $\pm 0.4$ \end{tabular} & \begin{tabular}{c} $110$\\ $\pm 8$ \end{tabular} & \begin{tabular}{c} $5.77$\\ $\pm 0.5$ \end{tabular} & \begin{tabular}{c} $216$\\ $\pm 4$ \end{tabular} & \begin{tabular}{c} $1.72$\\ $\pm 0.4$ \end{tabular} \\
    \hline
    Rosen & \begin{tabular}{c} $7.47$\\ $\pm 1$ \end{tabular} & \begin{tabular}{c} ${\bf 1.53}$\\ ${\pm 0.01}$ \end{tabular} & \begin{tabular}{c} $32.6$\\ $\pm 4$ \end{tabular} & \begin{tabular}{c} $4.22$\\ $\pm 0.5$ \end{tabular} & \begin{tabular}{c} $38.7$\\ $\pm 3$ \end{tabular} & \begin{tabular}{c} $5.10$\\ $\pm 0.6$ \end{tabular} & \begin{tabular}{c} $46.4$\\ $\pm 1$ \end{tabular} & \begin{tabular}{c} $2.09$\\ $\pm 0.3$ \end{tabular} \\
    \hline
    Branin & \begin{tabular}{c} $1.08$\\ $\pm 0.09$ \end{tabular} & \begin{tabular}{c} ${\bf 0.42}$\\ ${\pm 0.003}$ \end{tabular} & \begin{tabular}{c} $3.08$\\ $\pm 0.4$ \end{tabular} & \begin{tabular}{c} $1.13$\\ $\pm 0.09$ \end{tabular} & \begin{tabular}{c} $2.81$\\ $\pm 0.4$ \end{tabular} & \begin{tabular}{c} $1.24$\\ $\pm 0.2$ \end{tabular} & \begin{tabular}{c} $2.31$\\ $\pm 0$ \end{tabular} & \begin{tabular}{c} $0.582$\\ $\pm 0.02$ \end{tabular} \\
    \hline
    Shekel & \begin{tabular}{c} $-5.28$\\ $\pm 0.4$ \end{tabular} & \begin{tabular}{c} ${\bf -6.79}$\\ ${\pm 0.5}$ \end{tabular} & \begin{tabular}{c} $-0.701$\\ $\pm 0.07$ \end{tabular} & \begin{tabular}{c} $-2.2$\\ $\pm 0.2$ \end{tabular} & \begin{tabular}{c} $-0.77$\\ $\pm 0.05$ \end{tabular} & \begin{tabular}{c} $-1.82$\\ $\pm 0.01$ \end{tabular} & \begin{tabular}{c} $-1.63$\\ $\pm 0$ \end{tabular} & \begin{tabular}{c} $-1.63$\\ $\pm 0$ \end{tabular} \\
    \hline
    Hartmann & \begin{tabular}{c} $-2.99$\\ $\pm 0.008$ \end{tabular} & \begin{tabular}{c} ${\bf -3.03}$\\ ${\pm 0.003}$ \end{tabular} & \begin{tabular}{c} $-2.45$\\ $\pm 0.03$ \end{tabular} & \begin{tabular}{c} $-2.97$\\ $\pm 0.006$ \end{tabular} & \begin{tabular}{c} $-2.54$\\ $\pm 0.04$ \end{tabular} & \begin{tabular}{c} $-2.97$\\ $\pm 0.005$ \end{tabular} & \begin{tabular}{c} $-2.41$\\ $\pm 0.01$ \end{tabular} & \begin{tabular}{c} $-2.42$\\ $\pm 0.02$ \end{tabular} \\
    \hline
  \end{tabular}
  }
\end{center}
\end{table*}

\subsection{Hyperparameter Optimization}

In the second experiment, we assess the performance of the proposed method on the hyperparameter optimization of machine learning algorithms.
We experiment with the following three machine learning algorithms, that are with the low budget setting in many cases.
\begin{itemize}
  \item MLP
  \item CNN
  \item LightGBM~\cite{ke2017lightgbm}
\end{itemize}

Table \ref{tab:mlp_cnn_hyperparameter} shows the four hyperparameters of MLP and their respective search spaces.
The MLP consists of two fully-connected layers and SoftMax at the end.
We set the maximum number of epochs during training to 20, and the mini-batch size to 128.
We use the MNIST dataset that has $28 \times 28$ pixel grey-scale images of digits, each belonging to one of ten classes.
The MNIST dataset consists of $60,000$ training images and $10,000$ testing images.
In this experiment, we split the training images into the training dataset of $50,000$ images and the validation dataset of $10,000$ images.

The CNN consists of two convolutional layers with batch normalization and SoftMax at the end.
Each convolutional layer is followed by a $3 \times 3$ max-pooling layer.
The two convolutional layers are followed by two fully-connected layers with ReLU activation.
We use the same hyperparameters and search spaces used by the MLP problem above (Table \ref{tab:mlp_cnn_hyperparameter}).
We set the maximum number of epochs during training to 10, and the mini-batch size to 128.
We use the MNIST dataset and split like the MLP problem.

\begin{table}[tb]
  \begin{center}
    \begin{tabular}{c}

      \begin{minipage}{0.47\hsize}
        \begin{center}
          \caption{Details of four hyperparameters of MLP and CNN optimized on MNIST dataset.}
          \label{tab:mlp_cnn_hyperparameter}
          \begin{tabular}{|c|c|}
            \hline
            Hyperparameter & Search Space \\
            \hline \hline
            learning rate of SGD & $[0.001, 0.20]$ \\
            \hline
            momentum of SGD & $[0.80, 0.999]$ \\
            \hline
            num of hidden nodes & $[50, 500]$ \\
            \hline
            dropout rate & $[0.0, 0.8]$ \\
            \hline
          \end{tabular}
        \end{center}
      \end{minipage}

      \begin{minipage}{0.47\hsize}
        \begin{center}
          \caption{Details of four hyperparameters of LightGBM optimized on Breast Cancer Wisconsin dataset.}
          \label{tab:lightgbm_hyperparameter}
          \begin{tabular}{|c|c|}
            \hline
            Hyperparameter & Search Space \\
            \hline \hline
            learning rate & $[0.001, 0.10]$ \\
            \hline
            colsample bytree & $[0.1, 1.0]$ \\
            \hline
            reg lambda & $[0.0, 100.0]$ \\
            \hline
            max depth & $[2, 7]$ \\
            \hline
          \end{tabular}
        \end{center}
      \end{minipage}

    \end{tabular}
  \end{center}
\end{table}

Table \ref{tab:lightgbm_hyperparameter} shows the four hyperparameters of LightGBM and their respective search spaces.
We use the Breast Cancer Wisconsin dataset~\cite{Dua:2017} that consists of $569$ data instances.
In the experiment using this dataset, we use $80 \%$ of the data instances as the training dataset, and the evaluation value is calculated using $7$-fold cross validation.

\subsubsection{Experimental Setting}
We run 50 trials for each experiment, and we set the evaluation budget to $B = 20$ in each trial.
For all experiments, we use the misclassification rate on the validation dataset as the evaluation value.
For all the problems, we regard the integer-valued hyperparameters as continuous variables by using rounded integer values when evaluating.

\subsubsection{Results}
\begin{figure*}[tb]
\centering
  \begin{minipage}[t]{.47\textwidth}
    \centering
    \includegraphics[width=70mm]{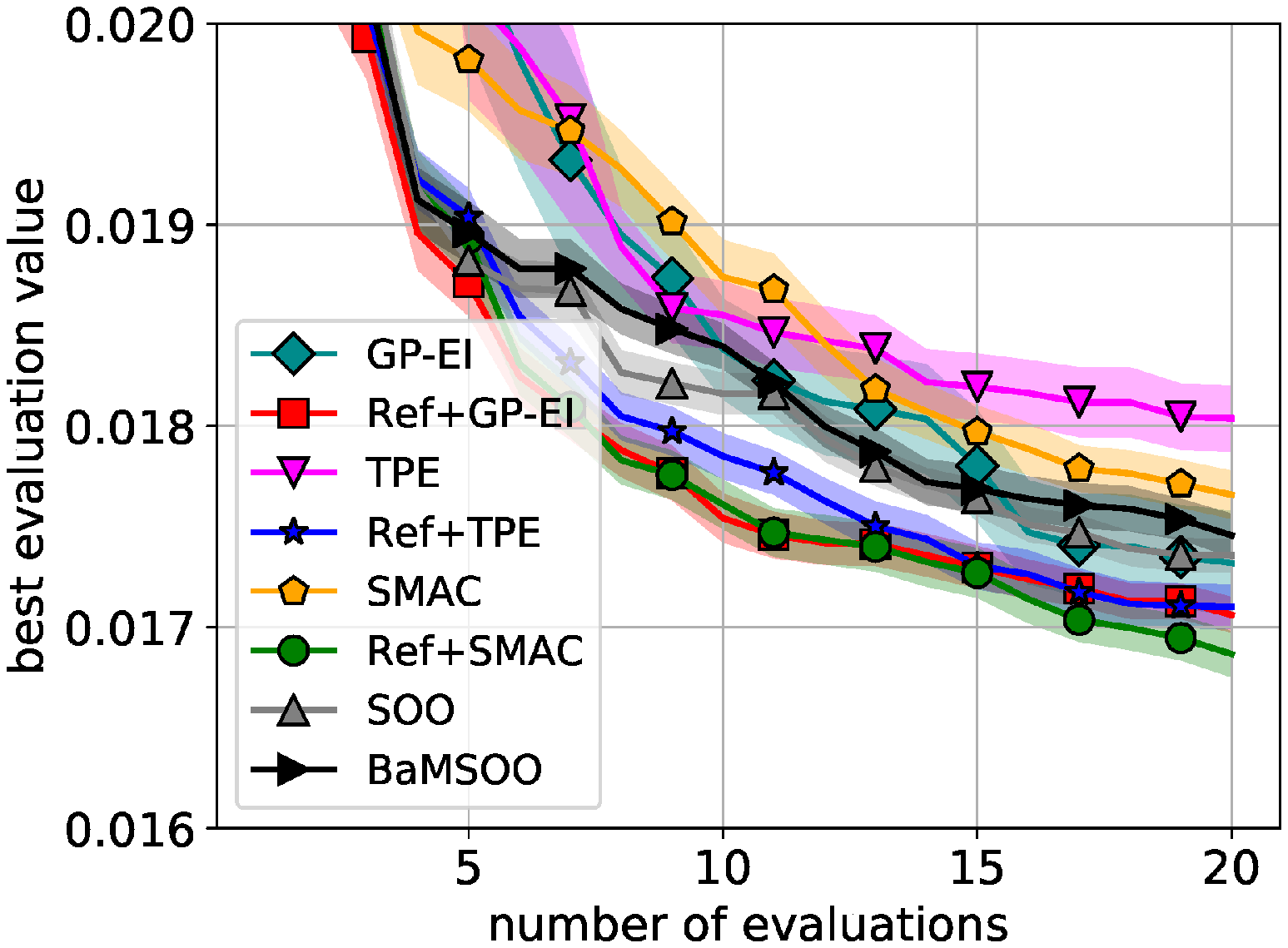}
    \subcaption{MLP with MNIST dataset}
    \label{fig:mean1}
  \end{minipage}
  \begin{minipage}[t]{.47\textwidth}
    \centering
    \includegraphics[width=70mm]{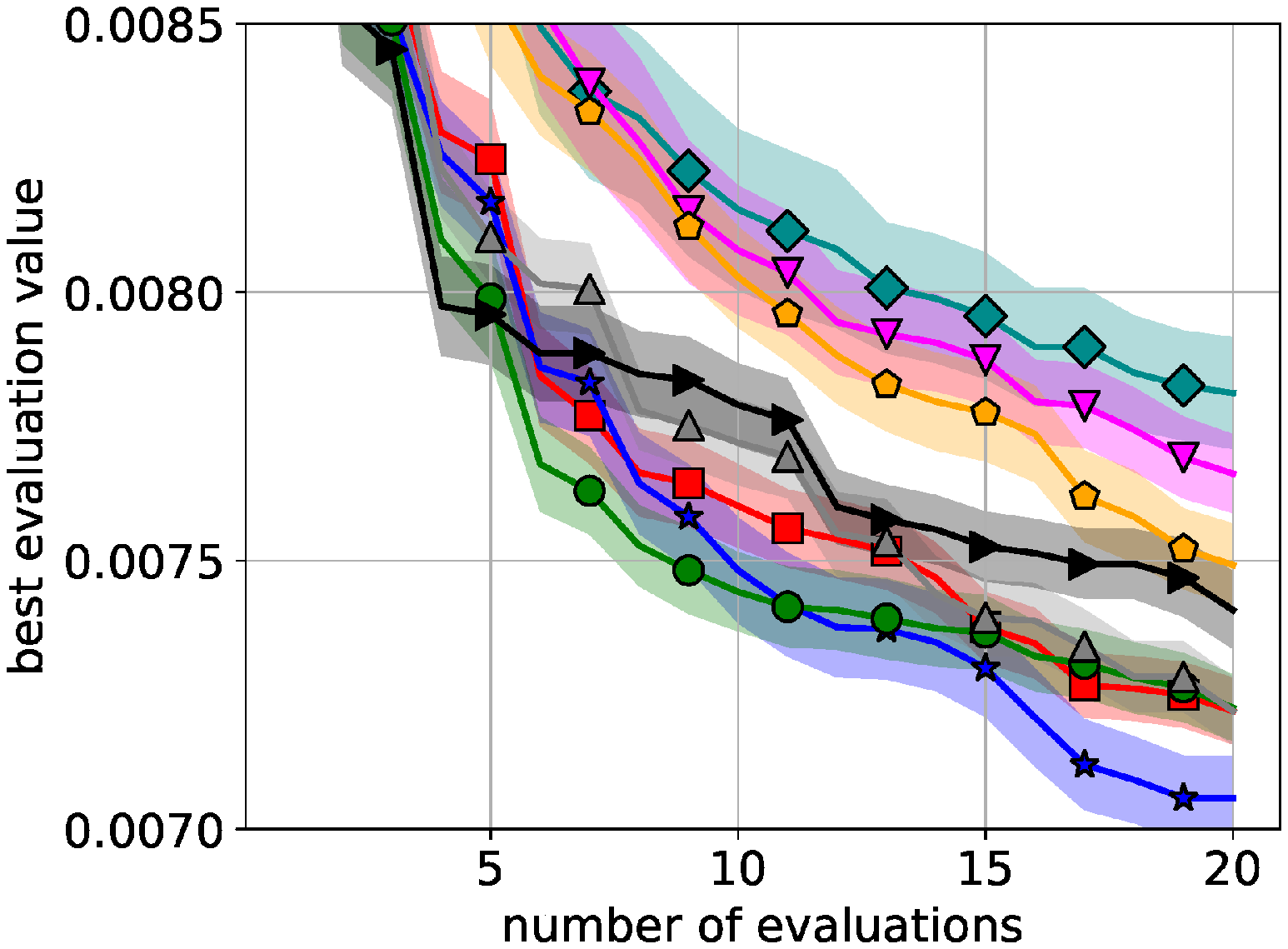}
    \subcaption{CNN with MNIST dataset}
    \label{fig:mean1}
  \end{minipage}
  \begin{minipage}[t]{.47\textwidth}
    \centering
    \includegraphics[width=70mm]{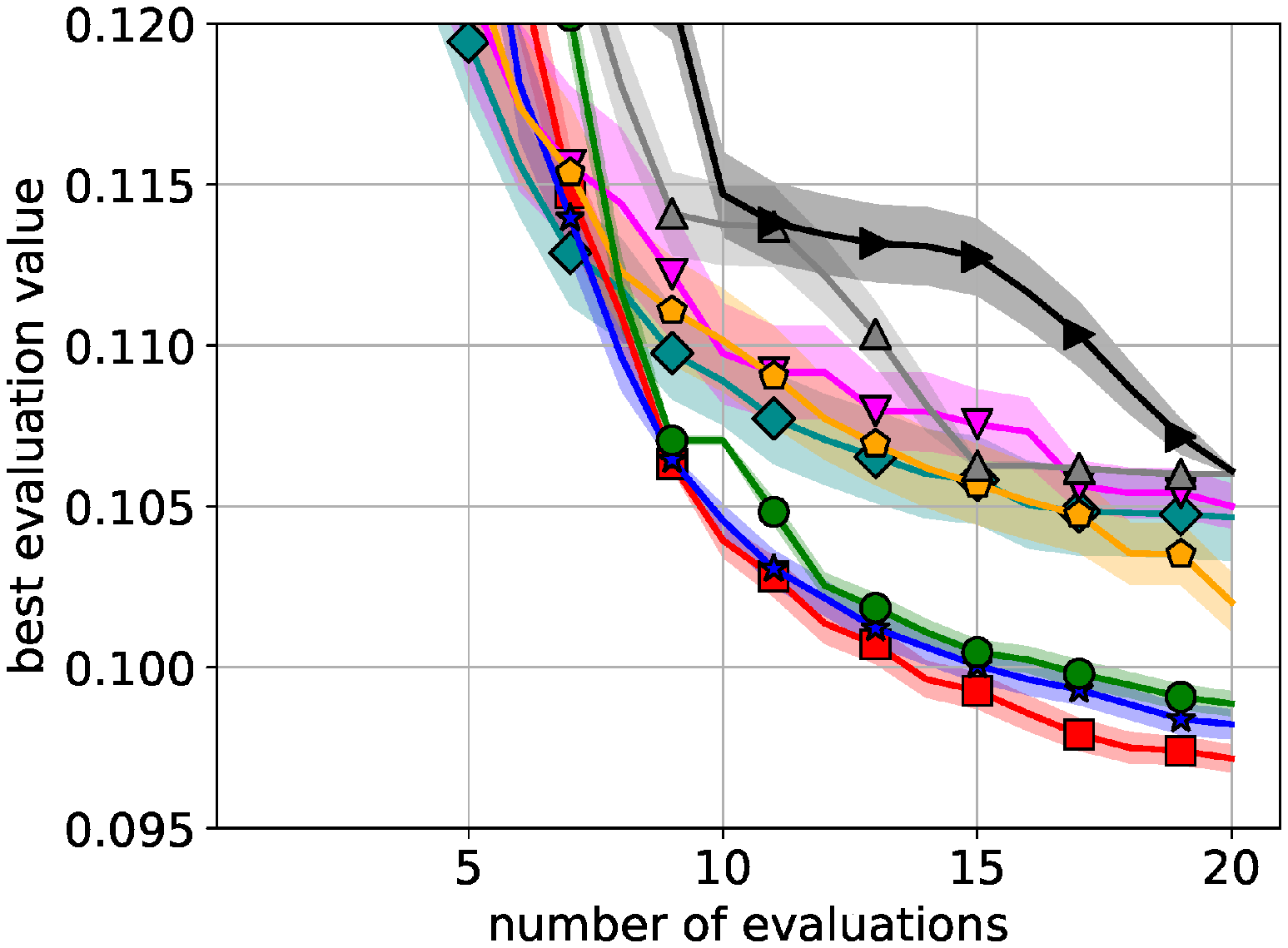}
    \subcaption{LightGBM with Breast Cancer Wisconsin dataset}
    \label{fig:mean1}
  \end{minipage}
  \caption{The sequences of the mean and standard error of the best evaluation values on the hyperparameter optimizations. The x-axis denotes the number of evaluations and the y-axis denotes the mean and standard error of the best evaluation values (averaged over 50 trials).}
  \label{fig:exp_hpo}
\end{figure*}

Figure \ref{fig:exp_hpo} and Table \ref{tab:hpo} show the mean and standard error of the best evaluation values for 50 trials on the hyperparameter optimization of the three machine learning algorithms.
Similar to the experiment of the benchmark functions, Ref+GP-EI, Ref+TPE and Ref+SMAC outperform GP-EI, TPE and SMAC in all the hyperparameter optimization of machine learning algorithms, respectively.
Likewise, Ref+GP-EI, Ref+TPE and Ref+SMAC show equal or better performance to SOO and BaMSOO in all problems.

\begin{table*}[tb]
\caption{Mean and standard error of the best evaluation values on the hyperparameter optimization of the machine learning algorithms. The bold line shows the best mean in all the methods. The values in each problem are multiplied by $10^2$ with the original values.}
\label{tab:hpo}
\small
\begin{center}
  {\tabcolsep = 1mm
  \begin{tabular}{c||c|c|c|c|c|c|c|c}
    \hline
    Problem & GP-EI & Ref+GP-EI & TPE & Ref+TPE & SMAC & Ref+SMAC & SOO & BaMSOO \\
    \hline \hline
    MLP & \begin{tabular}{c} $1.73$\\ $\pm 0.02$ \end{tabular} & \begin{tabular}{c} $1.71$\\ $\pm 0.008$ \end{tabular} & \begin{tabular}{c} $1.80$\\ $\pm 0.01$ \end{tabular} & \begin{tabular}{c} $1.71$\\ $\pm 0.01$ \end{tabular} & \begin{tabular}{c} $1.77$\\ $\pm 0.01$ \end{tabular} & \begin{tabular}{c} ${\bf 1.69}$\\ ${\pm 0.01}$ \end{tabular} & \begin{tabular}{c} $1.74$\\ $\pm 0.008$ \end{tabular} & \begin{tabular}{c} $1.75$\\ $\pm 0.01$ \end{tabular} \\
    \hline
    CNN & \begin{tabular}{c} $0.781$\\ $\pm 0.01$ \end{tabular} & \begin{tabular}{c} $0.722$\\ $\pm 0.006$ \end{tabular} & \begin{tabular}{c} $0.766$\\ $\pm 0.007$ \end{tabular} & \begin{tabular}{c} ${\bf 0.706}$\\ ${\pm 0.008}$ \end{tabular} & \begin{tabular}{c} $0.749$\\ $\pm 0.008$ \end{tabular} & \begin{tabular}{c} $0.723$\\ $\pm 0.006$ \end{tabular} & \begin{tabular}{c} $0.722$\\ $\pm 0.006$ \end{tabular} & \begin{tabular}{c} $0.741$\\ $\pm 0.007$ \end{tabular} \\
    \hline
    LightGBM & \begin{tabular}{c} $10.5$\\ $\pm 0.1$ \end{tabular} & \begin{tabular}{c} ${\bf 9.72}$\\ ${\pm 0.04}$ \end{tabular} & \begin{tabular}{c} $10.5$\\ $\pm 0.07$ \end{tabular} & \begin{tabular}{c} $9.82$\\ $\pm 0.05$ \end{tabular} & \begin{tabular}{c} $10.2$\\ $\pm 0.09$ \end{tabular} & \begin{tabular}{c} $9.89$\\ $\pm 0.04$ \end{tabular} & \begin{tabular}{c} $10.6$\\ $\pm 0.005$ \end{tabular} & \begin{tabular}{c} $10.6$\\ $\pm 0.01$ \end{tabular} \\
    \hline
  \end{tabular}
  }
\end{center}
\end{table*}

\section{Conclusion}
In this study, we developed a simple heuristic method for Bayesian optimization with the low budget setting.
The proposed method refines the promising region by dividing the region at equal intervals for each dimension.
By refining the search space, Bayesian optimization can be executed with a promising region as the initial search space.

We experimented with the six benchmark functions and the hyperparameter optimization of the three machine learning algorithms (MLP, CNN, LightGBM).
We confirmed that Bayesian optimization with the proposed method outperforms Bayesian optimization alone in all the problems including the benchmark functions and the hyperparameter optimization.
Likewise, Bayesian optimization with the proposed method shows equal or better performance to two search-space division algorithms.

In future work, we plan to adapt the proposed method for noisy environments.
Real-world problems such as hyperparameter optimization are often noisy; thus, making the optimization method robust is important.
Furthermore, because we do not consider the variable dependency at present, we are planning to refine the search space taking the variable dependency into consideration.

{\small
\bibliographystyle{unsrt}  
\bibliography{references}  
}

\end{document}